\documentclass[10pt,twocolumn,letterpaper]{article}

\usepackage{multirow}
\usepackage[pagenumbers]{wacv} 

\usepackage{appendix}

\definecolor{wacvblue}{rgb}{0.21,0.49,0.74}
\usepackage{xcolor}
\definecolor{brickred}{rgb}{0.8, 0.25, 0.33}
\definecolor{midnightblue}{rgb}{0.1, 0.1, 0.44}
\definecolor{oceanboatblue}{rgb}{0.0, 0.47, 0.75}

\usepackage{arydshln}
\usepackage[table,xcdraw]{xcolor}
\usepackage[pagebackref,breaklinks,colorlinks,allcolors=wacvblue]{hyperref}
\usepackage{pifont}

\def\wacvPaperID{325} 
\def\confName{WACV}
\def\confYear{2027}

\title{MaskMerge: Structure-aware Token Merging for Efficient Instance Segmentation}

\author{Bavesh Balaji\\
University Of Waterloo\\
{\tt\small bbalaji@uwaterloo.ca}
\and
Letian Jiang\\
University Of Waterloo\\
{\tt\small l236jiang@uwaterloo.ca}
\and
Alexander Wong\\
University Of Waterloo\\
{\tt\small alexander.wong@uwaterloo.ca}
\and
David Clausi\\
University Of Waterloo\\
{\tt\small dclausi@uwaterloo.ca}
\and
Sirisha Rambhatla\\
University Of Waterloo\\
{\tt\small sirisha.rambhatla@uwaterloo.ca}
}

\begin{document}
\maketitle
\begin{abstract}
Token merging improves the efficiency of Vision Transformers by combining redundant tokens, but existing methods largely rely on feature similarity to determine which tokens to merge. This criterion is effective for semantic segmentation, yet can fail in instance segmentation, where adjacent objects of the same class often have highly similar features. We identify this failure mode as \textbf{cross-instance token fusion}. We introduce \textbf{MaskMerge}, an instance-aware token merging method that uses coarse intermediate mask predictions to constrain feature-based merging and preserve instance boundaries. We evaluate MaskMerge on two diverse instance segmentation domains characterized by high inter-instance visual similarity: PanNuke and iSAID. Against the SOTA token merging method ALGM, MaskMerge improves mAP from 30.1 to 32.1 on PanNuke while reducing computation from 229 to 153 GFLOPs and running 1.18$\times$ faster, and from 33.1 to 36.2 on iSAID while reducing computation from 198 to 182 GFLOPs and running 1.23$\times$ faster. Additionally, we also test it on Cityscapes and match ALGM's performance while using fewer FLOPs. 
The gains also transfer to an MAE-pretrained backbone, demonstrating that instance-aware merging enables more aggressive compression when appearance similarity alone cannot reliably preserve object identity.

\end{abstract}
    
\section{Introduction}
\label{sec:intro}

Vision Transformers (ViTs) have become powerful backbones for dense visual recognition, including semantic \cite{strudel2021segmenter, xie2021segformer, zhang2022segvit}, instance \cite{kerssies2025eomt, cheng2021mask2former, liu2023revisiting}, and panoptic segmentation \cite{kirillov2019panoptic}. Their ability to model long-range interactions enables strong spatial reasoning, but comes at a substantial computational cost: self-attention scales quadratically with the number of visual tokens \cite{yin2022adavit, meng2021adavitadaptivevisiontransformers}. This cost is particularly pronounced in segmentation, where high-resolution inputs are required to preserve fine spatial detail \cite{ALGM, lu2023content}. At the same time, dense ViT representations contain considerable redundancy, motivating a growing body of work on reducing the number of tokens processed by the network \cite{apt}.

\begin{figure}[t]
    \centering
    \includegraphics[width=\columnwidth]{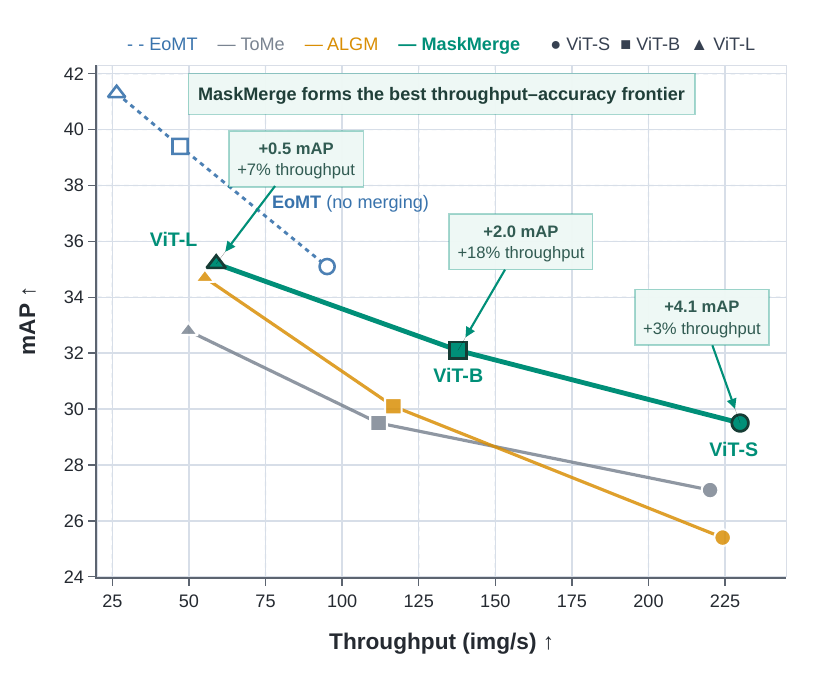}
    
    \vspace{-7px}
    \caption{\textbf{Throughput--accuracy Pareto frontier on PanNuke \cite{gamper2019pannuke}  across ViT-S, ViT-B, and ViT-L backbones.} MaskMerge consistently achieves a better accuracy--efficiency trade-off than ToMe \cite{ToMe} and ALGM \cite{ALGM} across model scales, indicating that instance-aware merging enables more aggressive token reduction while better preserving instance segmentation accuracy.}
    \label{fig:placeholder}
\end{figure}

Existing token reduction methods broadly fall into pruning, halting, and merging approaches. Token pruning removes representations deemed uninformative, but permanently discarding spatial locations can be undesirable for dense prediction \cite{ding2023prune, yu2023xpruner, li2022sait, wang2023zerotprune, yu2023xpruner, shang2024llavaprumerge}. Token halting reduces computation by terminating selected tokens at intermediate layers \cite{10483763, courdier2023paumer, tang2023dynamic}, but requires recovering their spatial information downstream. \emph{Token merging} instead combines redundant tokens while maintaining a correspondence to the original spatial support, making it particularly attractive for segmentation \cite{lu2023content, ToMe, ALGM, renggli2022tokenmereger, lee2024dtem}. Methods such as ToMe \cite{ToMe} and ALGM \cite{ALGM} have demonstrated that substantial computation can be removed with little or even no degradation on semantic segmentation.

\begin{figure*}[t]
    \centering

    \begin{subfigure}[c]{0.56\textwidth}
        \centering
        \includegraphics[width=\linewidth]{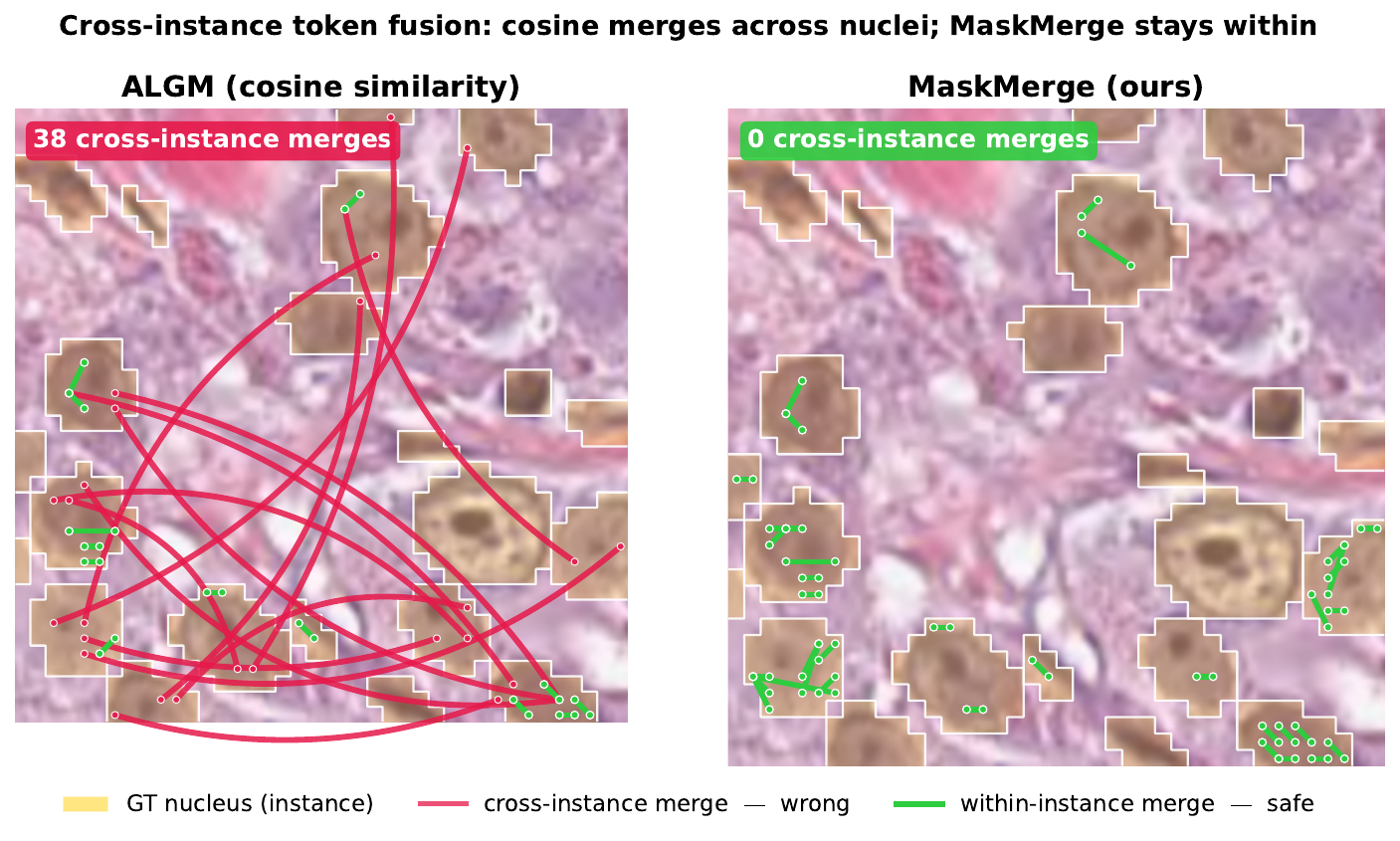}
        \caption{Qualitative viz. of cross-instance token fusion in ALGM.}
        \label{fig:cross_instance_qual}
    \end{subfigure}
    \hfill
    \begin{subfigure}[c]{0.41\textwidth}
        \centering
        \includegraphics[width=\linewidth]{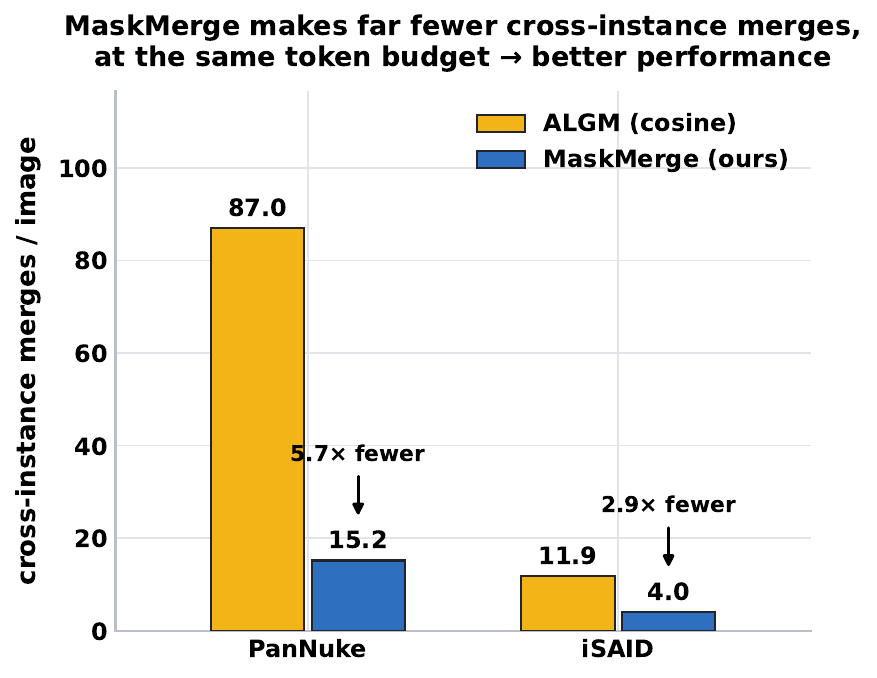}
        \caption{Cross-instance merge rate at a matched token budget.}
        \label{fig:cross_bar}
    \end{subfigure}

    \caption{
    \textbf{Cross-instance token fusion in similarity-based token merging.}
    \textbf{(a)} ALGM frequently merges tokens across different nuclei instances, whereas MaskMerge constrains merges within predicted instances.
    \textbf{(b)} At the same merge budget, ALGM produces $5.7\times$ and $2.9\times$ more cross-instance merges than MaskMerge on PanNuke \cite{gamper2019pannuke} and iSAID \cite{zamir2019isaid}, respectively, demonstrating that instance-aware constraints enable safer token reduction.
    }
    \label{fig:cross_instance_fusion}

    \vspace{-2mm}
\end{figure*}

A central assumption underlying most existing token merging methods is that \emph{feature similarity is a reliable proxy for merge compatibility} \cite{ToMe, ALGM, nguyen2026structsam}. Tokens with high cosine similarity are treated as redundant and are therefore natural candidates for aggregation. For semantic segmentation, this assumption is often well aligned with the task: neighboring regions representing the same semantic category should receive the same prediction, and merging similar features within such regions is generally benign.

The same assumption becomes considerably less reliable for \textbf{instance segmentation} specifically for homogeneous data. Here, the model must not only recognize what is present, but also preserve \emph{which object} each spatial location belongs to \cite{liu2018path}. Adjacent instances of the same category can have extremely similar visual representations despite requiring different output masks. Two neighboring nuclei in a histopathology image \cite{gamper2019pannuke} or vehicles parked side-by-side in an aerial image \cite{zamir2019isaid} may therefore be highly similar in feature space while belonging to different instances. A similarity-only merge criterion may consequently aggregate tokens across instance boundaries, as shown in Fig.~\ref{fig:cross_instance_qual}, a failure mode we term \textbf{cross-instance token fusion}. Fig.~\ref{fig:cross_bar} quantifies this effect: at the same merge budget, MaskMerge reduces cross-instace merges by $\mathbf{5.7\times}$ and $\mathbf{2.9\times}$ on PanNuke and iSAID respectively in comparison to ALGM \cite{ALGM}. This shows that feature similarity alone is insufficient to determine whether two tokens are safe to merge for an instance-level task, motivating the use of predicted instance structure to enable both safer and more aggressive token reduction.


Further, cross-instance token fusion affects not only accuracy, but also the attainable efficiency. Increasing the merge rate of a similarity-based method eventually introduces more erroneous cross-instance merges, forcing it to retain additional tokens to preserve segmentation quality. A safer merge criterion can instead support \emph{more aggressive compression} without destroying instance boundaries. Thus, instance awareness offers a dual benefit: it preserves object identity while enabling greater inference throughput.

These insights motivate \textbf{MaskMerge}, an instance-aware token merging framework that incorporates predicted mask structure directly into the merge decision. Modern query-based segmentation models \cite{cheng2021mask2former, cheng2021maskformer, kerssies2025eomt} progressively form object-specific representations throughout the network. MaskMerge exploits this intermediate structure rather than introducing an independent grouping mechanism. At an early stage of the encoder, a lightweight mask prediction operation combines the current object queries with the visual tokens to obtain coarse instance masks. These masks provide an inexpensive estimate of the instance assignment of each token. During subsequent token merging, MaskMerge restricts candidate matches to tokens assigned to the same predicted mask before selecting merges according to feature similarity. The resulting operation retains the efficiency and simplicity of similarity-based token merging while substantially reducing opportunities for cross-instance fusion.

We evaluate this distinction in two complementary domains where visually homogeneous instances make appearance-only merging particularly challenging: \textbf{PanNuke} \cite{gamper2019pannuke} for nuclei instance segmentation and \textbf{iSAID} \cite{zamir2019isaid} for aerial instance segmentation. Across these domains, the instance constraint enables MaskMerge to operate at substantially more aggressive compression levels without incurring the degradation observed with similarity-only merging. On PanNuke, MaskMerge reaches \textbf{32.1 mAP}, compared with 30.1 mAP for ALGM \cite{ALGM}, while reducing computation from 229 to 153 GFLOPs and increasing throughput from 121.5 to 142.8 images/s. On iSAID, it improves mAP from 33.1 to \textbf{36.2}, while reducing computation from 198 to 182 GFLOPs and increasing throughput from 168.8 to 206.9 tiles/s. Additionally, to test MaskMerge on real-world data, we ran ALGM and Maskmerge on Cityscapes, both methods achieve 19.8 AP, but MaskMerge requires only 298 GFLOPs compared with 347 GFLOPs. MaskMerge outperforms ALGM \cite{ToMe} and ToMe \cite{ToMe} \textbf{across different backbone scales} on PanNuke as shown in Fig.~\ref{fig:placeholder}, demonstrating a better accuracy-throughput tradeoff than cosine methods. 
Moreover, the benefit is not tied to a single initialization: replacing the standard backbone with an MAE-pretrained variant on PanNuke preserves the advantage, with MaskMerge obtaining 34.0 mAP compared with 32.0 mAP for Compact ALGM.

Together, these results suggest a broader principle for efficient instance-level recognition: \emph{visual redundancy and task-level redundancy are not equivalent}. Two tokens may be nearly indistinguishable in appearance yet encode different information for an instance segmentation objective. Efficient token reduction should therefore consider not only \emph{how similar} two representations are, but also \emph{whether the task permits them to share a representation}. MaskMerge provides a lightweight realization of this principle by using the model's emerging instance structure to guide token aggregation.

Our contributions are threefold:
\begin{itemize}
    \item We identify and characterize \textbf{cross-instance token fusion}, a failure mode of similarity-based token merging in instance segmentation. Through per-instance analysis, we show that its impact is concentrated in crowded regions containing adjacent, visually similar instances.
    
    \item We introduce \textbf{MaskMerge}, an instance-aware token merging method that uses intermediate coarse masks to constrain feature-based aggregation. MaskMerge exploits structural information already present in the segmentation model and requires neither additional supervision nor an auxiliary training objective.
    
    \item We demonstrate a consistently improved \textbf{accuracy--efficiency trade-off} across three distinct instance segmentation domains. MaskMerge outperforms similarity-based ALGM on PanNuke and iSAID and matches its accuracy on Cityscapes at substantially lower computation. The improvement also transfers to a backbone with a different pre-training strategy, supporting the robustness of the proposed merge criterion.
\end{itemize}

\section{Related Works}
\subsection{ViTs for Instance Segmentation}

Vision Transformers (ViTs) are well suited to instance segmentation because self-attention captures global object context, while hierarchical or multi-scale features preserve the spatial detail needed for precise mask prediction ~\cite{dosovitskiy2020vit,wang2021pyramid,li2022exploring}. Query-based approaches such as DETR represent objects as a set of learned queries, enabling separate predictions for multiple instances of the same category without handcrafted proposal generation~\cite{zhu2020deformable, carion2020endtoend, liu2021swin, liu2021swinv2}. Building on this, MaskFormer reframed segmentation as mask classification, providing a unified formulation that supports multiple segmentation tasks~\cite{cheng2021maskformer}. Mask2Former extended this paradigm with masked attention, improving both accuracy and scalability for universal image segmentation~\cite{cheng2021mask2former}. More recently, EoMT showed that generic ViT backbones can act as strong universal segmentation models when equipped with suitable token-to-mask prediction heads, highlighting the potential of model-agnostic ViT architectures for unified segmentation~\cite{kerssies2025eomt}. Our work follows this line by enhancing efficiency within ViT-based segmentation frameworks while preserving instance structure.

\subsubsection{Token reduction in general.}
In image classification and related settings, token reduction is mainly achieved by pruning, merging, or hybrid strategies. Pruning methods remove uninformative tokens using auxiliary predictors or attention-derived saliency scores, and are often optimized to preserve top-layer semantics under reduced token budgets \cite{rao2021dynamicvitefficientvisiontransformers, kong2022spvit, meng2021adavitadaptivevisiontransformers, fayyaz2022ats, liang2022evit, wang2023zerotprune, yin2022adavit, yu2023xpruner, yang2023global, zheng2022savit}. Beyond token dropping, efficiency is also improved through compressed token representations and quantization-aware formulations \cite{wu2018compressed, park2023rgb, liu2021post, li2022q, moon2024instance}, as well as through alternative attention mechanisms such as linearized kernels and local-window computation \cite{katharopoulos2020transformers, lu2021soft, liu2021swin, wei2023sparsifiner, chen2023sparsevit}. 

Merging approaches reduce sequence length by combining redundant tokens either with learned mappings \cite{zong2022selfslimmed, renggli2022tokenmereger, ryoo2021tokenlearner, chang2023making} or with similarity-driven matching rules \cite{ToMe, heo2023local_global, bonnaerens2023learned_threshold, chen2023diffrate, cao2023pumer, long2022beyond, wei2023joint, liang2022evit, shang2024llavaprumerge, tran2024accelerating, kallini2024mrt5, lee2024dtem}. In practice, these methods differ in when reduction is applied (early vs.\ layer-wise), whether token identities remain recoverable, and how much extra routing computation is introduced. Hybrid methods explicitly combine pruning, merging, and/or token fusing to improve the speed--accuracy frontier \cite{bonnaerens2023learned_threshold, chen2023diffrate, cao2023pumer, long2022beyond, wu2023ppt}. Despite strong gains, many approaches still rely on learned token-selection policies or threshold schedulers, which can increase optimization complexity, reduce hardware-friendliness, and limit end-to-end training-time acceleration \cite{meng2021adavitadaptivevisiontransformers, yin2022adavit, choudhury2025rlt, ding2023prune}.

\subsubsection{Token reduction for semantic segmentation.}
Applying token reduction to semantic segmentation is more constrained because dense prediction requires reconstructing per-location outputs while preserving object boundaries and fine structures. Classification-oriented methods can be adapted by tracking assignment maps and remapping merged tokens to original positions \cite{chang2023making, ToMe, liang2022evit, heo2023local_global, long2022beyond}, but direct transfer is often brittle and can degrade mIoU, especially around thin regions and small objects \cite{liang2022expediting}. This creates a stronger requirement for spatially faithful reduction mechanisms than in global classification.

Task-specific segmentation approaches therefore emphasize recoverability. Token pausing/halting methods skip updates for confident regions and restore them later \cite{courdier2023paumer, tang2023dynamic, 10483763}, improving efficiency but reducing cross-region interaction once tokens are paused. Clustering-based methods such as ELViT and AiluRus merge neighboring redundant tokens while maintaining dense outputs \cite{liang2022expediting, li2023ailurus}; however, practical gains are constrained by clustering overhead, one-shot grouping, and limited adaptability to scene-level global redundancy. CTS introduces a CNN policy for token sharing before the transformer \cite{lu2023content}, which improves locality-aware reduction but adds model complexity. ALGM \cite{ALGM} improves upon previous token merging methods by adding a local module and a global merging module inspired by ToMe \cite{ToMe}. While these methods have been great for semantic segmentation, they have not been evaluated for instance segmentation. Additionally, prior work focuses on using feature cosine similarity as the metric to merge tokens. However, for instance-level tasks, this leads to many erroneous merges which deterioirates their performance in instance segmentation. Hence, we incorporate an additional object-level mask constraint to mitigate these erroneous merges and improve performance on these tasks.
\section{Methodology}
\label{sec:method}

\begin{figure*}[t]
    \centering
    \includegraphics[width=0.9\linewidth]{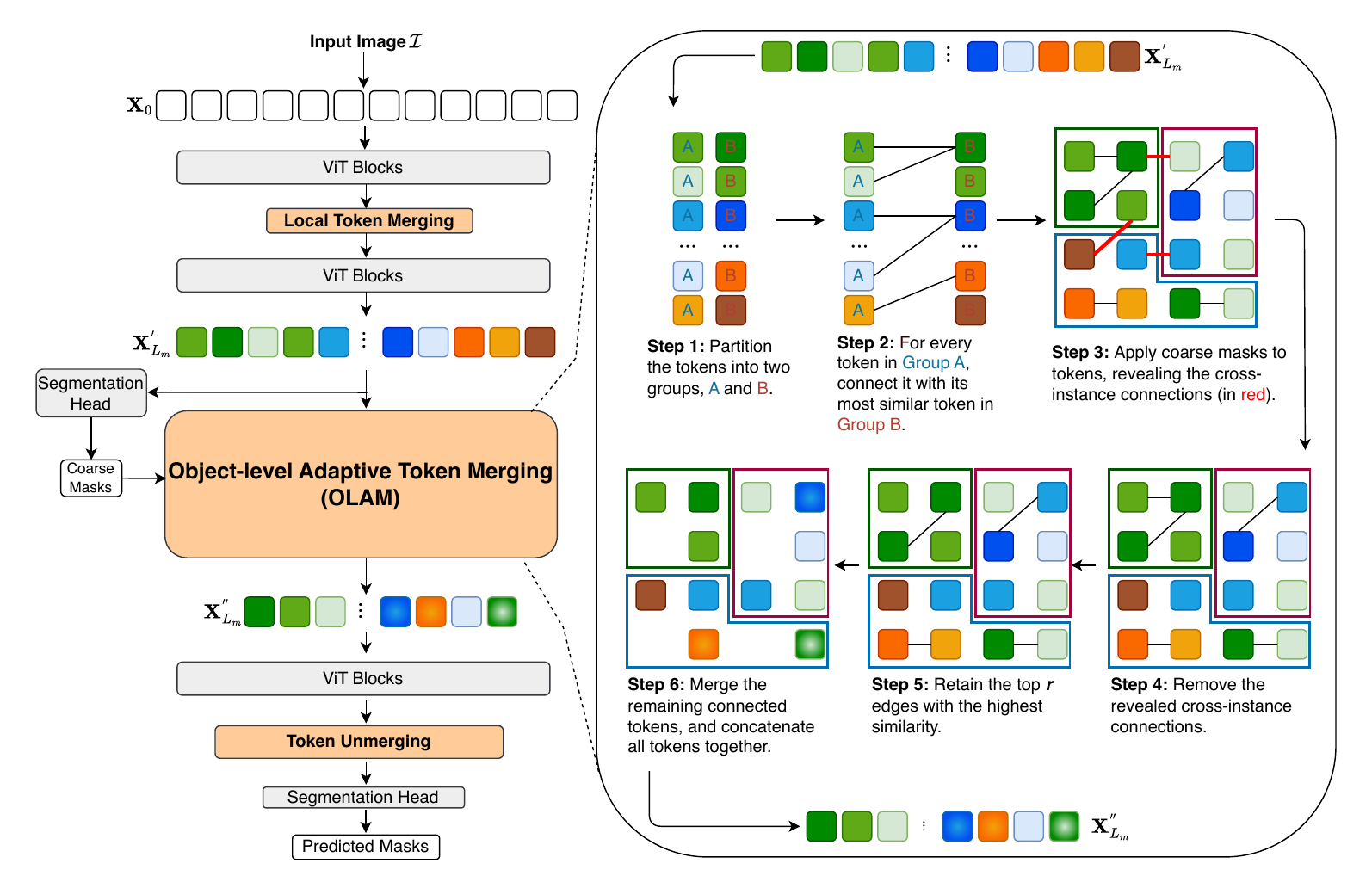}
    \caption{\textbf{Overall pipeline of MaskMerge}. Given an input image $\mathbf{I} \in \mathbb{R}^{H \times W}$, the standard patch tokens $\mathbf{X}_0 \in \mathbb{R}^{N \times D}$ of the image are initially passed through $L_1$ layers of the transformer encoder. After that, similar tokens within a small neighbourhood are merged together using the local merge module resulting in $\mathbf{X}_{L_1}' \in \mathbb{N' \times D}$ where $N' < N$. After this, the model is passed through $L_m$ transformer layers to get $\textbf{X}_{L_m}'$ where our object-level adaptive token merging (OLAM) module further reduces tokens from $N'$ to $N''$ while preserving instance identities. These reduced tokens are prcoessed by the rest of the transformer encoder layers and then unmerged back to the original dimensions before being fed to the segmentation head. }
    \label{fig:main}
\end{figure*}

In this section, we introduce MaskMerge, our \textbf{mask-based adaptive token merging framework} for efficient universal image segmentation. Our approach leverages object-level mask proposals as proxies to define our token merging metric, enabling the model to perform segmentation and token refinement jointly within a unified transformer architecture. We first revisit the standard patch-based tokenization used in Vision Transformers ~\cite{dosovitskiy2020vit} (Sec.~\ref{sec:patch}). We then present our mask-based token merging module in Sec.~\ref{sec:seg_method}. The overall architecture is illustrated in Fig. \ref{fig:main}.

\subsection{Vision Transformers with Patch-based Tokenization}
\label{sec:patch}

Vision Transformers (ViTs) represent an image \( I \in \mathbb{R}^{3 \times H \times W} \) as a sequence of tokens obtained from fixed-size patches. 
Specifically, the image is partitioned into \(N = \tfrac{HW}{p^2}\) non-overlapping patches of size \(p \times p\). 
Each patch is flattened and projected through a linear embedding layer to produce a token representation of dimension \(D\). 
This yields the initial token sequence
\[
\mathbf{X}_0 = \{\mathbf{x}_1, \dots, \mathbf{x}_N\} \in \mathbb{R}^{D \times N},
\]
where \(\mathbf{x}_i \in \mathbb{R}^D\) denotes the embedding of the \(i\)-th patch.

The tokens are processed through a stack of \(L\) Transformer blocks. 
Each block consists of a Multi-Head Self-Attention (MHSA) module followed by a two-layer feed-forward network (MLP), both preceded by Layer Normalization. 
The update rule at layer \(i\) is given by
\begin{equation}
\begin{aligned}
&\mathbf{Z}_i = \mathbf{X}_i + \mathrm{MHSA}(\mathrm{Norm}(\mathbf{X}_i)), \\
&\mathbf{X}_{i+1} = \mathbf{Z}_i + \mathrm{MLP}(\mathrm{Norm}(\mathbf{Z}_i)).
\end{aligned}
\label{eq:vit}
\end{equation}

After the final layer, the token sequence \(\mathbf{X}_L\) can be reshaped into a spatial feature map
$
F_{\text{ViT}} \in \mathbb{R}^{D \times \frac{H}{p} \times \frac{W}{p}}.
$

While patch-based tokenization provides a structured representation, it lacks spatial adaptivity. All image regions contribute an equal number of tokens regardless of their semantic complexity. Consequently, simple regions incur redundant computation while complex object boundaries remain under-represented. To address this limitation, we introduce an \emph{adaptive token merging} mechanism that dynamically merges tokens while preserving object-level structure.

\subsection{Mask-based Token Merging}
\label{sec:seg_method}

To introduce spatial adaptivity and reduce redundancy, we progressively transform fixed patch tokens into content-aware tokens through a two-stage merging strategy: 
(i) \textit{local token merging} to remove redundancy within small spatial neighborhoods, and 
(ii) \textit{object-level adaptive merging} that consolidates tokens belonging to the same instance.

\subsubsection{Local Token Merging}

Given the token sequence after the first \(L_1\) transformer layers,
\[
\mathbf{X}_{L_1} = \{\mathbf{x}_1, \dots, \mathbf{x}_N\} \in \mathbb{R}^{D \times N},
\]
we perform an initial merging step to remove redundant tokens in local spatial neighborhoods following \cite{ALGM}.

We first partition the tokens into non-overlapping windows of size \(W \times W\) in the spatial grid induced by the original patch layout. 
Let \(\mathcal{G}_k\) denote the set of tokens within the \(k\)-th window and \(n_k\) be the number of tokens in that window. 
For each window, we compute the average pairwise cosine similarity
\begin{equation}
\mu_k = 
\frac{2}{n_k(n_k-1)} 
\sum_{i<j,\, i,j \in \mathcal{G}_k}
\frac{\mathbf{x}_i^\top \mathbf{x}_j}
{\|\mathbf{x}_i\|_2 \|\mathbf{x}_j\|_2}.
\end{equation}

If the similarity exceeds a threshold \(\tau\), the tokens in the window are merged into a single token. 
The merged token representation is computed using size-aware averaging
\begin{equation}
\mathbf{x}_k^{\text{merge}} =
\frac{1}{n_k}
\sum_{i \in \mathcal{G}_k} \mathbf{x}_i.
\end{equation}

Formally, for $\mathbf{X}_{L_1}' \in \mathbb{R}^{D \times N'}, \quad N' < N$, the merging operation defines a mapping
\[
\mathcal{M}_{\text{local}} : \mathbf{X}_{L_1} \rightarrow \mathbf{X}_{L_1}'.
\]

This step reduces token redundancy while preserving spatial locality, which has been shown to be important for stable early-layer representations \cite{ALGM, lu2023content}.

\subsubsection{Object-level Adaptive Token Merging (OLAM)}

While local merging removes redundant tokens within small neighborhoods, it does not account for higher-level object structure. 
We therefore introduce a global merging mechanism that leverages instance-level information predicted by the segmentation model.

After passing the locally merged tokens through an additional \(L_m\) transformer layers, we obtain
\[
\mathbf{X}_{L_m}' =
\{\mathbf{x}_i\}_{i=1}^{N'} \in \mathbb{R}^{D \times N'}.
\]

We construct a bipartite matching graph between two partitions of the token set like \cite{ToMe, ALGM}. 
Let the tokens be divided into two disjoint subsets
\[
A = \{\mathbf{x}_i\}_{i=1}^{N'/2}, \quad
B = \{\mathbf{x}_j\}_{j=N'/2+1}^{N'}.
\]

The similarity between tokens is computed as
\begin{equation}
S_{ij} =
\frac{\mathbf{x}_i^\top \mathbf{x}_j}
{\|\mathbf{x}_i\|_2 \|\mathbf{x}_j\|_2}.
\end{equation}

However, relying solely on feature similarity can result in merges across different object instances. 
To address this limitation, we incorporate object-consistency constraints using coarse instance masks predicted by the segmentation model.

Let $\mathcal{M}^c = \{M_q^c\}_{q=1}^{Q}$
denote the set of coarse masks associated with \(Q\) object queries. 
Each token is assigned an object identifier
\begin{equation}
o_i =
\arg\max_{q \in \{1,\dots,Q\}} M_q^c(i),
\end{equation}
where \(M_q^c(i)\) denotes the mask value at the spatial location corresponding to token \(i\). We then define a mask-constrained similarity as 
\begin{equation}
\tilde{S}_{ij} =
\begin{cases}
S_{ij}, & o_i = o_j, \\
-\infty, & \text{otherwise}.
\end{cases}
\end{equation}

To further improve robustness, we incorporate a mask confidence-aware constraint. 
Let
\[
c_i = \max_q M_q^c(i)
\]
denote the confidence of predicted mask corresponding to token \(i\). 
We allow merging only if
\begin{equation}
\min(c_i, c_j) \ge \tau_\ell,
\end{equation}
where \(\tau_\ell\) is a training-dependent threshold that gradually decreases as training progresses. Given the masked similarity matrix \(\tilde{S}\), we compute bipartite matching pairs and merge each pair using weighted averaging
\begin{equation}
\mathbf{x}_{ij}^{\text{merge}} =
\frac{w_i \mathbf{x}_i + w_j \mathbf{x}_j}{w_i + w_j},
\end{equation}
where \(w_i, w_j\) denote token sizes that track how many original patches contributed to each token. This produces the final merged token sequence
\[
\mathbf{X}_{L_m}'' \in \mathbb{R}^{D \times N''}, \quad N'' < N'.
\]

\subsubsection{Token Unmerging}

After the OLAM module, the merged tokens \(\mathbf{X}_{L_m}''\) are processed by the remaining transformer layers. 
Before passing the features to the segmentation decoder, the tokens must be restored to their original spatial resolution. To enable this, we maintain a lightweight index mapping during merging. 
Specifically, for each merged token \(k\), we store the set of original token indices
\[
\mathcal{I}_k = \{i_1, i_2, \dots, i_m\},
\]
representing all tokens that contributed to the merged representation.

During the unmerging step, the embedding of token \(k\) is copied back to all indices in \(\mathcal{I}_k\):
\begin{equation}
\mathbf{x}_{i}^{\text{unmerge}} = \mathbf{x}_k,
\quad \forall i \in \mathcal{I}_k.
\end{equation}

This restores the token sequence to its original length \(N\), allowing the decoder to operate on a dense spatial grid.

Unlike prior methods that maintain dense source matrices of size \(N \times N\) to track token ancestry \cite{ToMe}, our approach maintains only a compact index list for each merged token. 
This sparse bookkeeping mechanism reduces memory overhead from \(O(N^2)\) to \(O(N)\) while enabling efficient reconstruction of the original token layout.

\section{Experiments}
\label{sec:experiments}

We evaluate MaskMerge on three questions:
(1) does instance-aware merging improve the accuracy--efficiency trade-off over similarity-based merging across domains with high inter-instance similarity?
(2) does the benefit generalize across backbone pre-training strategies? and
(3) can the same principle be applied to a segmentation architecture in which mask prediction is decoupled from the backbone?

\subsection{Experimental Setup}

\paragraph{Datasets.}
We evaluate on three instance segmentation datasets spanning substantially different visual domains. \textbf{PanNuke} \cite{gamper2019pannuke} contains nuclei from histopathology images across five classes; we train on folds 1--2 and evaluate on fold 3. \textbf{iSAID} \cite{zamir2019isaid} contains 15 categories in high-resolution aerial imagery and includes many small, visually similar object instances; we use $512\times512$ tiles. Additionally, to test it on real-world data, we use \textbf{Cityscapes} \cite{cordts2016cityscapes}. It  contains eight instance categories in urban driving scenes, where visually similar vehicles and pedestrians frequently occur in close proximity. We evaluate on the 500-image validation set using $512\times1024$ sliding-window crops. \

\paragraph{Model and training.}
Our primary experiments use EoMT with a DINOv2 ViT-B backbone. Following ALGM, we apply local merging in the first encoder block and perform instance-aware merging at layer 5. For the PanNuke ViT-S and ViT-L experiment, we applied global merge at layers 5 and 8 respectively. We initialize from the trained model and fine-tune with token merging enabled for 15 epochs to make sure that the mask queries are adapted to the intermediate representations.

We compare against ToMe \cite{ToMe} and ALGM \cite{ALGM} as feature-similarity-based token merging baselines. We additionally evaluate an MAE-pretrained ViT-B backbone to study sensitivity to backbone pre-training, and apply MaskMerge to Mask2Former with a DINOv2 ViT-B backbone to examine whether instance-aware merging transfers beyond EoMT.

\paragraph{Metrics.}
We report instance segmentation mean average precision (mAP), end-to-end throughput, and GFLOPs. Throughput and FLOPs are measured with batch size 4 using FP16 forward inference. Throughput is reported in images/s for PanNuke and tiles/s for Cityscapes and iSAID.

\subsection{Main Results}

Table~\ref{tab:main_results} compares MaskMerge with standard full-token EoMT inference and two similarity-based token merging baselines. ToMe and ALGM determine merge candidates from feature similarity, whereas MaskMerge additionally constrains candidate pairs using predicted instance structure.

\begin{table}[t]
\centering
\caption{\textbf{Instance segmentation across datasets and backbone scales.}
All models use EoMT with DINOv2-pretrained ViTs. Throughput is images/s for PanNuke and tiles/s for iSAID and Cityscapes.}
\label{tab:main_results}

\resizebox{\linewidth}{!}{
\begin{tabular}{lllccc}
\toprule
Dataset & Backbone & Method & mAP $\uparrow$ & Throughput $\uparrow$ & GFLOPs $\downarrow$ \\
\midrule

\multirow{12}{*}{PanNuke}

& \multirow{4}{*}{ViT-S}
& EoMT \cite{kerssies2025eomt} & 35.1 & 95.1 & 140.8 \\
& & ToMe \cite{ToMe} & 27.1 & 220.1 & 76.1 \\
& & ALGM \cite{ALGM} & 25.4 & 224.2 & 41.6 \\
& & \textbf{MaskMerge} & \textbf{29.5} & \textbf{229.9} & \textbf{35.3} \\

\cmidrule(lr){2-6}

& \multirow{4}{*}{ViT-B}
& EoMT \cite{kerssies2025eomt} & 39.4 & 49.1 & 623 \\
& & ToMe \cite{ToMe} & 29.5 & 114 & 369 \\
& & ALGM \cite{ALGM} & 30.1 & 121.5 & 229 \\
& & \textbf{MaskMerge} & \textbf{32.1} & \textbf{142.8} & \textbf{153} \\

\cmidrule(lr){2-6}

& \multirow{4}{*}{ViT-L}
& EoMT \cite{kerssies2025eomt} & 41.3 & 26.4 & 1654 \\
& & ToMe \cite{ToMe} & 32.8 & 49.8 & 1093 \\
& & ALGM \cite{ALGM} & 34.7 & 55.2 & 546 \\
& & \textbf{MaskMerge} & \textbf{35.2} & \textbf{58.9} & \textbf{482} \\

\midrule

\multirow{4}{*}{iSAID}
& \multirow{4}{*}{ViT-B}
& EoMT \cite{kerssies2025eomt} & 41.6 & 83.9 & 383 \\
& & ToMe \cite{ToMe} & 31.5 & 160.5 & 207 \\
& & ALGM \cite{ALGM} & 33.1 & 168.8 & 198 \\
& & \textbf{MaskMerge} & \textbf{36.2} & \textbf{206.9} & \textbf{182} \\

\midrule

\multirow{4}{*}{Cityscapes}
& \multirow{4}{*}{ViT-B}
& EoMT \cite{kerssies2025eomt} & 21.6 & 39.5 & 827 \\
& & ToMe \cite{ToMe} & 10.6 & \textbf{109.7} & 330 \\
& & ALGM \cite{ALGM} & 19.8 & 107.4 & 347 \\
& & \textbf{MaskMerge} & \textbf{19.8} & 105.2 & \textbf{298} \\

\bottomrule
\end{tabular}
}
\end{table}

MaskMerge consistently improves the accuracy--efficiency trade-off over similarity-based merging across datasets and backbone scales. On PanNuke, it improves over ALGM by \textbf{2.0 mAP} on ViT-B while using $1.50\times$ fewer FLOPs and achieving $1.18\times$ higher throughput. Fig.~\ref{fig:placeholder} and Ta. ~\ref{tab:main_results} further shows gains of \textbf{4.1 mAP} and \textbf{0.5 mAP} on ViT-S and ViT-L, with corresponding throughput improvements of 3\% and 7\%. On iSAID, MaskMerge improves mAP by \textbf{3.1 points}, reduces computation by $1.08\times$, and runs $1.23\times$ faster. We additionally evaluate on the more heterogeneous real-world Cityscapes benchmark, where MaskMerge matches ALGM at 19.8 mAP while using fewer FLOPs. Its throughput is similar to ALGM because coarse-mask prediction introduces runtime overhead; on the more homogeneous PanNuke and iSAID datasets, however, high inter-instance similarity allows MaskMerge to perform substantially more aggressive yet instance-consistent merging, and the resulting token reduction more than offsets this overhead.

The results highlight an important limitation of similarity-based merging. As the merge rate increases, feature similarity alone can increasingly combine visually similar tokens from different instances. ALGM must therefore retain more tokens to avoid substantial accuracy degradation. By constraining candidate merges with predicted instance structure, MaskMerge supports more aggressive token reduction while better preserving instance separation. This effect is particularly pronounced on PanNuke and iSAID, where distinct instances frequently exhibit highly similar local appearance.

MaskMerge also consistently outperforms ToMe despite operating at comparable or lower computational cost. On PanNuke, ToMe requires 369 GFLOPs yet reaches only 29.5 mAP, compared with 32.1 mAP at 153 GFLOPs for MaskMerge. Similarly, on iSAID, MaskMerge improves mAP from 31.5 to 36.2 while using fewer FLOPs. These results suggest that increasing the aggressiveness of appearance-only merging cannot provide the same accuracy--efficiency trade-off as incorporating instance structure into the merge decision.

\subsection{Generalization Across Backbone Pre-training}

We next test whether the benefit of MaskMerge depends on the DINOv2 representation. We replace it with an MAE-pretrained ViT-B backbone and repeat the PanNuke experiment.

\begin{table}[t]
\centering
\caption{\textbf{Backbone generalization on PanNuke.} Results use an MAE-pretrained ViT-B backbone.}
\label{tab:mae}
\begin{tabular}{lcc}
\toprule
Method & mAP $\uparrow$ & AP$_\mathrm{S}$ $\uparrow$ \\
\midrule
EoMT \cite{kerssies2025eomt} & 36.8 & 36.0 \\
ALGM \cite{ALGM} & 32.0 & 31.5 \\
\textbf{MaskMerge} & \textbf{34.0} & \textbf{33.4} \\
\bottomrule
\end{tabular}
\end{table}

MaskMerge improves overall mAP from 32.0 to \textbf{34.0} and small-instance performance from 31.5 to \textbf{33.4} as shown in Tab.~\ref{tab:mae}. The advantage therefore persists under a substantially different self-supervised pre-training strategy, suggesting that the improvement is not specific to the DINOv2 feature representation.

\subsection{Extension to Mask2Former}

EoMT exposes intermediate query--token interactions directly in the encoder, making coarse instance masks readily available for MaskMerge. We next study whether the same principle can be applied to \textbf{Mask2Former} \cite{cheng2021mask2former}, where mask prediction occurs in a separate decoder.

A direct implementation requires two passes: a dense forward pass to obtain decoder masks followed by a second mask-guided pass. We therefore additionally introduce a lightweight single-pass variant in which a two-layer convolutional head predicts coarse instance structure directly from backbone features. The head learns from the decoder masks of the frozen model and is used at inference time to constrain token merging.

\begin{table}[t]
\centering
\caption{\textbf{MaskMerge with Mask2Former on PanNuke.}
All methods use the same DINOv2 ViT-B backbone. The single-pass variant predicts coarse instance structure directly from backbone features, while the two-pass variant obtains masks from a separate dense forward pass.}
\label{tab:mask2former}
\resizebox{\linewidth}{!}{
\begin{tabular}{lcccc}
\toprule
Method & Passes & mAP $\uparrow$ & GFLOPs $\downarrow$ & img/s $\uparrow$ \\
\midrule
Mask2Former \cite{cheng2021mask2former} & 1 & 33.89 & 167.9 & 36.1 \\
ALGM \cite{ALGM} & 1 & 24.44 & 154.7 & 36.6 \\
ToMe \cite{ToMe} & 1 & 30.69 & 158.3 & 36.1 \\
\textbf{MaskMerge} & \textbf{1} & \textbf{32.53} & \textbf{161.2} & \textbf{36.1} \\
\bottomrule
\end{tabular}}
\end{table}

The results in Table~\ref{tab:mask2former} show that instance-aware merging remains beneficial when mask prediction is decoupled from the backbone. The single-pass MaskMerge retains 32.53 mAP, only 1.36 points below the standard Mask2Former model, compared with 30.69 for ToMe and 24.44 for ALGM. This indicates that predicted instance structure provides a substantially safer criterion for token aggregation than feature similarity alone.


These results demonstrate that the principle underlying MaskMerge is not specific to EoMT. When intermediate instance masks are unavailable, a lightweight predictor can provide sufficient structural information to guide token aggregation in a decoupled segmentation architecture. The particularly favorable efficiency gains observed with EoMT arise because instance information is directly available within the model, whereas Mask2Former requires an auxiliary prediction mechanism.

\subsection{Ablation Study}
In order to understand more about the different components of our technique, we conduct rigorous ablations and highlight the importance of our design choices. All the ablations are conducted on PanNuke using EoMT-B.

\subsubsection{Impact of each module} Table~\ref{tab:ablation_component} isolates the contribution of OLAM from conventional cosine-similarity-based global merging. Using global cosine merging alone gives 29.5 mAP, whereas replacing the similarity-only criterion with OLAM improves performance to 33.4 mAP, showing that predicted instance structure provides a substantially safer merge signal. The difference is even clearer when combined with local merging: cosine-based global merging reaches 30.1 mAP at 229 GFLOPs, while OLAM attains 32.1 mAP with only 153 GFLOPs and increases throughput from 121.5 to 142.8 images/s. This indicates that OLAM is better suited to aggressive early token reduction, as its instance-aware constraint allows more tokens to be merged without sacrificing instance separation.

\begin{table}[t]
\centering
\caption{Impact of OLAM module position on overall performance on PanNuke. The \colorbox{oceanboatblue!20}{blue} row indicates the final setting used for our main table result.}
\label{tab:OLAM_pos}

\begin{small}
\begin{tabular}{cccc}
\toprule
Layer & mAP$\uparrow$ & GFLOPs$\downarrow$ & img/s$\uparrow$ \\
\midrule
4 & 31.4 & 145 & 153.15 \\
\cellcolor{oceanboatblue!20}5
& \cellcolor{oceanboatblue!20}32.1
& \cellcolor{oceanboatblue!20}153
& \cellcolor{oceanboatblue!20}142.8 \\
6 & 33.0 & 171 & 134.2 \\
7 & 33.8 & 181 & 125.6 \\
\bottomrule
\end{tabular}
\end{small}

\vspace{-0.1in}
\end{table}

\begin{table}[t]
\centering
\caption{Impact of the local and OLAM modules on overall performance on PanNuke. The \colorbox{oceanboatblue!20}{blue} row indicates the final setting used for our main table result.}
\label{tab:ablation_component}

\resizebox{\columnwidth}{!}{
\begin{tabular}{ccc|ccc}
\toprule
Local & OLAM & OLAM w/o mask & mAP$\uparrow$ & GFLOPs$\downarrow$ & img/s$\uparrow$ \\
\midrule
\ding{55} & \ding{55} & \ding{51} & 29.5 & 369 & 114 \\
\ding{51} & \ding{55} & \ding{51} & 30.1 & 229 & 121.5 \\
\ding{55} & \ding{51} & \ding{55} & 33.4 & 287 & 116.2 \\
\rowcolor{oceanboatblue!20}
\ding{51} & \ding{51} & \ding{55} & 32.1 & 153 & 142.8 \\
\bottomrule
\end{tabular}
}

\vspace{-0.1in}
\end{table}

\subsubsection{OLAM module position} Table~\ref{tab:OLAM_pos} studies the trade-off induced by the layer at which OLAM is applied. Applying OLAM earlier yields the highest efficiency, but the coarse instance predictions are less mature and lead to lower accuracy; placing it at layer 4 reaches 31.4 mAP at 145 GFLOPs. Delaying OLAM progressively improves accuracy, reaching 33.8 mAP at layer 7, but retains more tokens for longer and therefore increases computation and reduces throughput. We choose layer 5 as the default operating point, which achieves 32.1 mAP at 153 GFLOPs and 142.8 images/s, providing a favorable balance between accuracy and efficiency.

\section{Conclusion}

We introduced \textbf{MaskMerge}, an instance-aware token merging framework that addresses \emph{cross-instance token fusion}, a key failure mode of similarity-based merging for instance segmentation. By using intermediate coarse masks to constrain token aggregation, MaskMerge preserves instance structure while enabling more aggressive token reduction. This leads to consistently better accuracy--efficiency trade-offs than cosine-based merging on PanNuke and iSAID, where visually similar instances are common, and the advantage persists across ViT scales and different backbone pre-training strategies. MaskMerge also matches ALGM on the more heterogeneous Cityscapes benchmark at lower computation and transfers to Mask2Former. Overall, our results show that effective token reduction for instance segmentation should account not only for feature similarity, but also for \emph{instance identity}, making structural guidance a simple and effective principle for efficient dense prediction.
{
    \small
    \bibliographystyle{ieeenat_fullname}
    \bibliography{main}
}

\end{document}


\begin{center}
    {\LARGE \textbf{Supplementary Material}} \\[0.5em]
    {\large \textbf{MaskMerge: Structure-aware Token Merging for Efficient Instance
Segmentation}}
\end{center}

\vspace{1em}

\appendix

This appendix contains additional material about the implementation details of MaskMerge in \ref{sec:imp}, additional ablations \ref{sec:abl}, and qualitative results in \ref{sec:qual}.
\vspace{-2.5mm}
\section{Implementation Details}
\label{sec:imp}

As mentioned in Sec. 4.1, we use the EoMT \cite{kerssies2025eomt} model predominantly with DINOv2-S/B/L backbones to investigate token merging on homogeneous domains. We initially train the EoMT model on each dataset (PanNuke \cite{gamper2019pannuke}, iSAID \cite{zamir2019isaid} and Cityscapes \cite{cordts2016cityscapes}) for 60 epochs from DINOv2's checkpoint and randomly initialized values for the other learnable weights. We use AdamW as our optimizer with a learning rate of 1e-4. We use LR warmup for 1000 steps and then decay the learning to 2e-5. After training the base model, to adapt the instance queries to the intermediate representation of the DINOv2 backbone, we further finetune the base model MaskMerge's token merging enabled for 15 epochs. Specifically, we re-enable mask annealing that is used in EoMT to make sure the training dynamocs are stable. Also, because the instance queries initially do not provide precise coarse masks with intermediate layer representations, we use confidence-based filtering to ensure that only confidence coarse masks  are allowed for merging. This makes sure that the mask predictions that are confident are allowed to merge initially, disallowing bad merges to happen and training to progress steadily, while the confidence is gradually reduced over the entire fine-tune period. Finally, at the end of fine-tuning, we have a model that can merge masks without any confidence-based filtering. We use an initial confidence of $\tau_{start}=0.7$ and end with $\tau_{end}=0$ gradually annealed over the 15 epochs. More analysis on this is presented in Tab. \ref{tab:annealing}. For finetuning, we restart with a new AdamW optimizer and a learning rate of 5e-5 and decay it to 2e-5. 

\section{Additional Ablations} \label{sec:abl}

We conducted additional ablations to study different aspects of training and design choices. Specifically, we conducted an experiment to find out the impact of confidence annealing on MAskMerge's OLAM module and the impact of computing coarse masks directly from locally merged tokens in the OLAM module instead of unmerging the there and then remerging again. This is important to study as unmerging before coarse mask computation improves performance but reduces throughput and increases FLOPs. All the ablations are conducted on PanNuke \cite{gamper2019pannuke} with EoMT-B model.

\begin{table}[t]
\centering
\caption{Impact of mask confidence annealing on PanNuke. The \colorbox{oceanboatblue!20}{blue} row indicates the final setting used for our main table result.}
\label{tab:annealing}

\begin{small}
\begin{tabular}{ccccc}
\toprule
$\tau_\ell^{\text{start}}$ & $\tau_\ell^{\text{end}}$ & mAP$\uparrow$ & GFLOPs$\downarrow$ & img/s$\uparrow$ \\
\midrule
0 & 0 & 30.4 & 153 & 142.8 \\
0.95 & 0.95 & {37.6} & 592 & 49.4 \\
0.95 & 0 & 32.0 & 153 & 142.8 \\
\rowcolor{oceanboatblue!20}
0.7 & 0 & {32.1} & {153} & {142.8} \\
0.5 & 0 & 31.6 & 153 & 142.8 \\
\bottomrule
\end{tabular}
\end{small}

\end{table}

\begin{table}[t]
\centering
\caption{Impact of computing the prediction head on locally merged tokens on PanNuke. The \colorbox{oceanboatblue!20}{blue} row indicates the final setting used for our main table result.}
\label{tab:prediction_head}

\begin{small}
\begin{tabular}{lccc}
\toprule
Token Type & mAP$\uparrow$ & GFLOPs$\downarrow$ & img/s$\uparrow$ \\
\midrule
Unmerged tokens & 32.8 & 199 & 74 \\
\rowcolor{oceanboatblue!20}
Locally merged tokens & 32.1 & {153} & {142.8} \\
\bottomrule
\end{tabular}
\end{small}

\end{table}

\begin{figure*}[t]
    \centering
    \includegraphics[width=\linewidth]{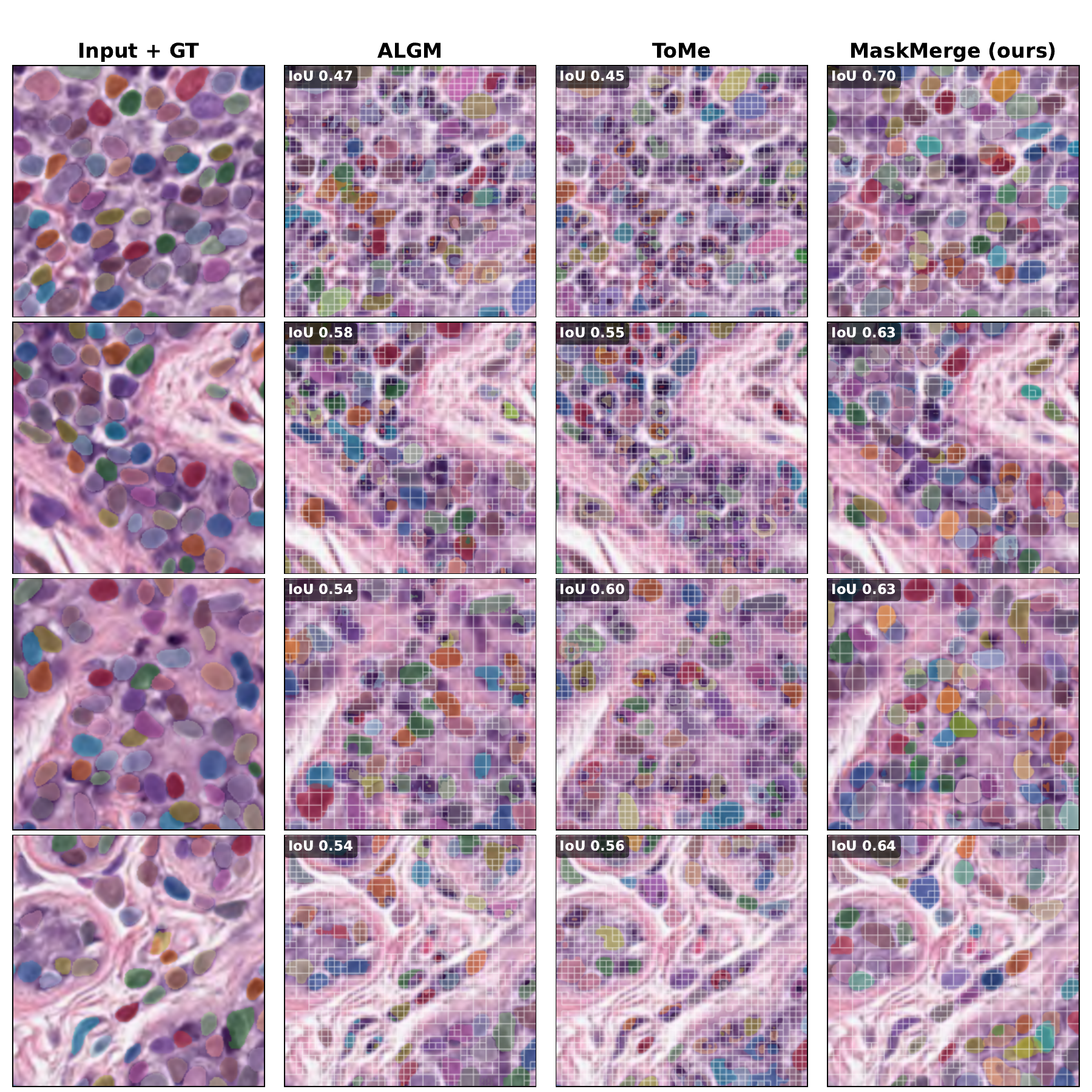}
    \caption{\textbf{Qualitative Results on PanNuke}. The GT and predicted masks for each image are overlaid on top of the input image and the token merges for each method is shown as white grids. Also, the IoU for each example with the GT is mentioned in the image itself. We can clearly see that MaskMerge promotes safe instance-aware merges that improve its IoU wrt cosine-based merging \cite{ALGM, ToMe}.}
    \label{fig:qual_pan}
\end{figure*}

\subsubsection{Impact of mask confidence annealing} Table~\ref{tab:annealing} studies the effect of confidence-threshold annealing for the coarse mask predictions. Without confidence filtering ($\tau_\ell^{\text{start}}=\tau_\ell^{\text{end}}=0$), performance drops to 30.4 mAP, indicating that unreliable early mask predictions can lead to harmful merge decisions. Keeping a high threshold throughout training improves accuracy substantially to 37.6 mAP, but prevents aggressive token reduction, increasing the cost to 592 GFLOPs and reducing throughput to 49.4 images/s. Annealing the threshold to zero provides a better accuracy--efficiency trade-off: starting from 0.7 achieves the best result of 32.1 mAP while retaining the full efficiency at 153 GFLOPs and 142.8 images/s. We therefore use $\tau_\ell^{\text{start}}=0.7$ and $\tau_\ell^{\text{end}}=0$ in all main experiments.

\begin{figure*}[t]
    \centering
    \includegraphics[width=\linewidth]{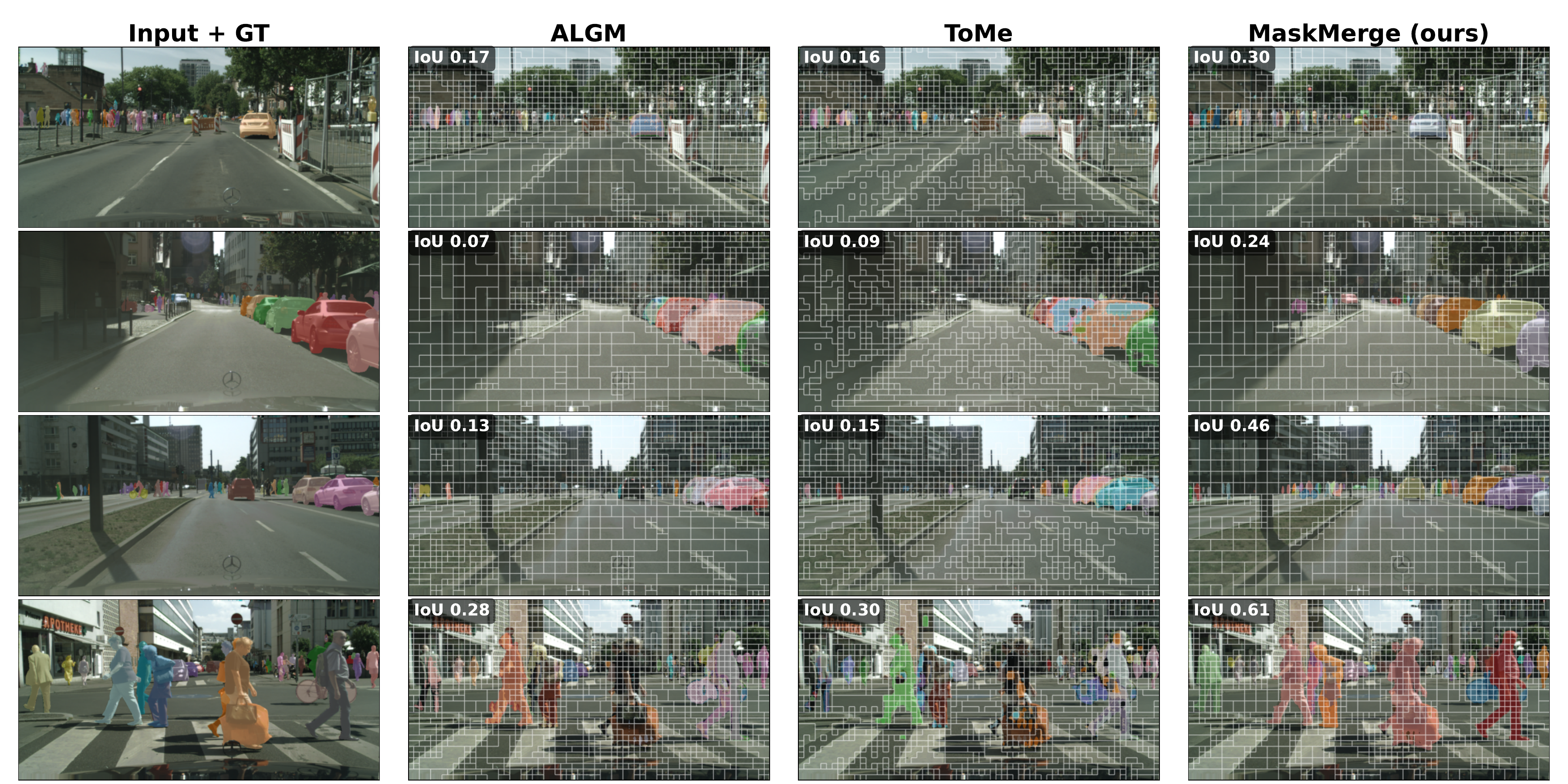}
    \caption{\textbf{Qualitative Results on Cityscapes}. The GT and predicted masks for each image are overlaid on top of the input image and the token merges for each method is shown as white grids. Also, the IoU for each example with the GT is mentioned in the image itself. Because of cityscapes' heterogeneous nature, we can see that MaskMerge does not provide that big of an improvement but still promotes safer merges than cosine-based merging \cite{ALGM, ToMe}.}
    \label{fig:qual_city}
\end{figure*}

\subsubsection{How does the number of tokens affect the intermediate mask prediction?} Table~\ref{tab:prediction_head} evaluates where the coarse mask prediction head is applied. Computing the masks from unmerged tokens slightly improves accuracy to 32.8 mAP, but incurs a substantial efficiency cost, requiring 199 GFLOPs and reducing throughput to 74 images/s. In contrast, predicting the coarse masks directly from the locally merged tokens retains nearly the same accuracy (32.1 mAP) while reducing computation to 153 GFLOPs and almost doubling throughput to 142.8 images/s. This design is therefore critical to making MaskMerge practical: it allows the instance-aware merge constraint to be computed from an already compressed representation, preserving the benefits of coarse structural guidance without negating the efficiency gained from token reduction.

\section{Qualitative Results}
\label{sec:qual}

\begin{figure*}
    \centering
    \includegraphics[width=\linewidth]{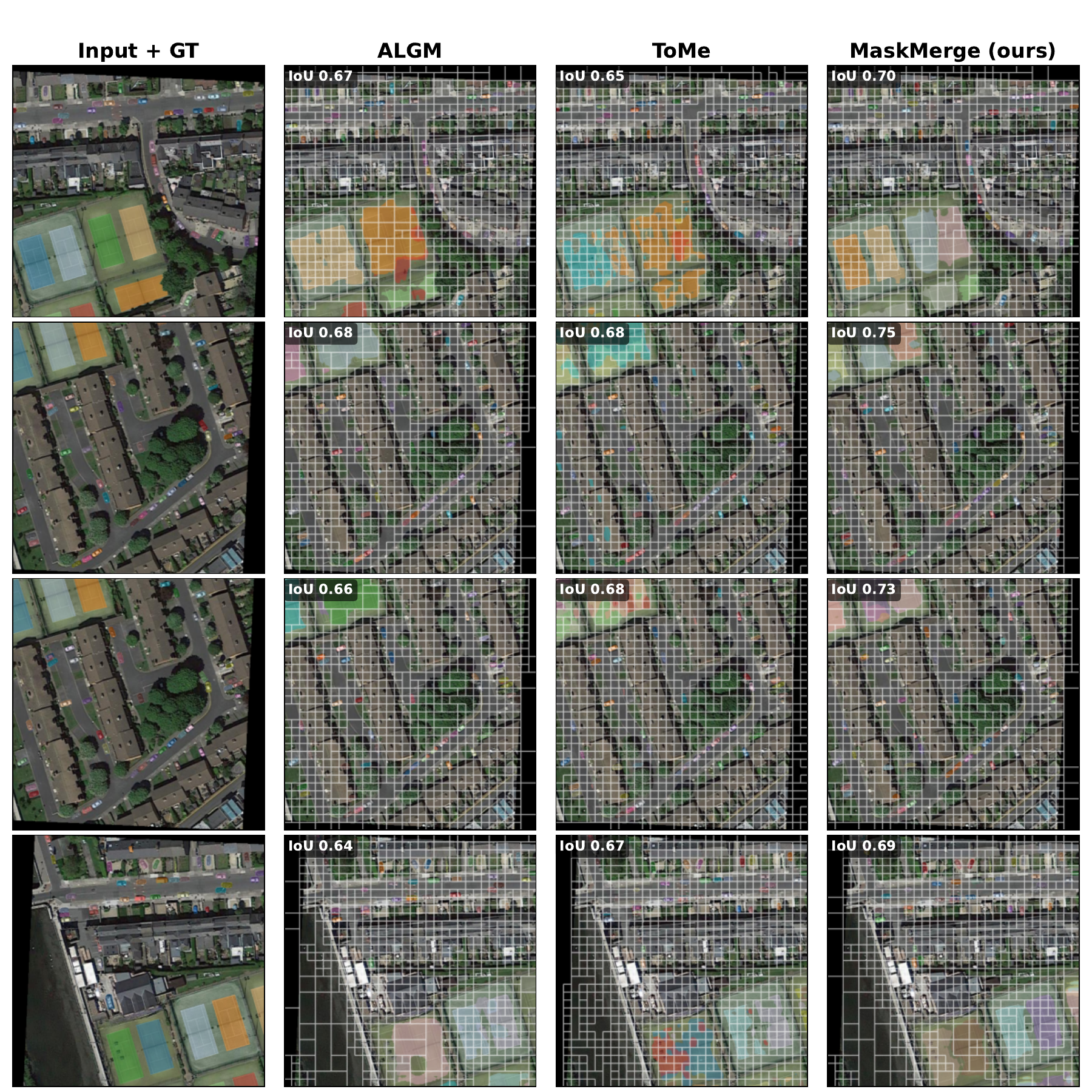}
    \caption{\textbf{Qualitative Results on iSAID}. The GT and predicted masks for each image are overlaid on top of the input image and the token merges for each method is shown as white grids. Also, the IoU for each example with the GT is mentioned in the image itself. We can clearly see that MaskMerge promotes safe instance-aware merges that improve its IoU wrt cosine-based merging \cite{ALGM, ToMe}.}
    \label{fig:qual_isaid}
\end{figure*}

In order to assess the significance of MaskMerge and understand the type of merges each method makes on different domains, we present qualitative results on all datasets in Figs. \ref{fig:qual_pan}, \ref{fig:qual_city} and  \ref{fig:qual_isaid}. In each image, the predicted masks of each method along with the resulting token merges are showcased. Additionally, the IoU of each method with the ground truth is also mentioned in the figures to indicate the performance improvement of MaskMerge. From the Figures, it is evident that MAskMerge is superior to ALGM \cite{ALGM} and ToMe \cite{ToMe}, signifying the limitation of appearance-driven merging in instance-level tasks and highlighting the need for instance-aware token reduction. 

{
    \small
    \bibliographystyle{ieeenat_fullname}
    \bibliography{main}
}